\documentclass[a4paper]{article}
\pdfoutput=1

\usepackage{INTERSPEECH2019,multirow,hyperref}
\usepackage{graphicx}
\usepackage{subcaption}

\newcommand{\scell}[2][c]{%
  \begin{tabular}[#1]{@{}c@{}}#2\end{tabular}}

\title{Constrained Output Embeddings for End-to-End Code-Switching Speech Recognition with Only Monolingual Data}

\name{Yerbolat Khassanov$^1$, Haihua Xu$^{2}$, Van Tung Pham$^{1,2}$, Zhiping Zeng$^2$,  Eng Siong Chng$^{1,2}$,\\ Chongjia Ni$^3$ and Bin Ma$^3$}
\address{
  $^1$School of Computer Science and Engineering, Nanyang Technological University, Singapore\\
  $^2$Temasek Laboratories, Nanyang Technological University, Singapore\\
  $^3$Machine Intelligence Technology, Alibaba Group}
\email{\{yerbolat002,haihuaxu,vantung001,zengzp,aseschng\}@ntu.edu.sg, \{ni.chongjia,b.ma\}@alibaba-inc.com}

\begin{document}

\maketitle
\begin{abstract}
The lack of code-switch training data is one of the major concerns in the development of end-to-end code-switching automatic speech recognition (ASR) models. 
In this work, we propose a method to train an improved end-to-end code-switching ASR using only monolingual data.
Our method encourages the distributions of output token embeddings of monolingual languages to be similar, and hence, promotes the ASR model to easily code-switch between languages.
Specifically, we propose to use Jensen-Shannon divergence and cosine distance based constraints. 
The former will enforce output embeddings of monolingual languages to possess similar distributions, while the later simply brings the centroids of two distributions to be close to each other. 
Experimental results demonstrate high effectiveness of the proposed method, yielding up to $4.5\%$ absolute mixed error rate improvement on Mandarin-English code-switching ASR task.

\ifx
  The lack of code-switch training data is one of the major concerns in the development of end-to-end code-switching ASR models.
  In this work, we propose a method to train an improved end-to-end code-switching ASR using only monolingual data.
  %Our method addresses the output token embedding representations of CS languages which disperse into two disjoint clusters due to the absence of CS training utterances, and hence, preventing the ASR model from switching between languages.
  Our method encourages the distributions of output token embeddings of monolingual languages to be similar, and hence, promotes the ASR model to easily code-switch between languages.
  Specifically, we propose to use Jensen-Shannon divergence and cosine distance based constraints.
  The former will enforce output embeddings of monolingual languages to possess similar distributions, while the later simply brings the centroids of two distributions to be close to each other.
  Experimental results demonstrate high effectiveness of the proposed method yielding up to $4.5\%$ absolute mixed error rate reduction on Mandarin-English code-switching ASR task.

  In this work, we aim to build end-to-end code-switching automatic speech recognition (E2E-CS-ASR) system using only monolingual data.
  While greatly alleviating the code-switch data scarcity problem, the E2E-CS-ASR will fail to learn language switch-points between and within utterances due to the absence of code-switching training samples.
  Consequently, the learned output representations of code-switching languages will diverge into separate clusters, hindering the E2E-CS-ASR from switching between languages.
  To address this problem, we propose to impose additional constraints which will force learned output representations of code-switching languages to be similar.
  Specifically, we propose to employ Jensen-Shannon divergence and cosine distance based constraints.
  The former one will enforce learned output representations to possess similar distributions, while the later one will enforce disjoint clusters to be close to each other.
  The experiment results performed on Mandarin-English code-switching language pair from the SEAME corpus demonstrate high effectiveness of the proposed method. 

  In spite of recent progress in code-switching speech recognition, the lack of code-switch data still remains a major challenge.
  %for many low-resource language pairs.
  %Prior works address this issue by using the semi-supervised and transfer learning based approaches which highly rely on the availability of code-mixed data.
  %On the contrary, we attempt to build an end-to-end code-mixed automatic speech recognition (ASR) system using only monolingual data for low-resource language pairs.
  Different from the previous works which highly rely on the availability of code-switch data, we aim to build an end-to-end code-switching automatic speech recognition (E2E-CS-ASR) system using only monolingual data.
  %While greatly simplifying the code-mixed ASR building pipeline, the end-to-end system will fail to learn language switch-points due to the absence of cross-lingual signal.
  While greatly mitigating the code-switch data scarcity problem, the E2E-CS-ASR will fail to learn language switch-points due to the absence of cross-lingual signal.
  Indeed, we investigate the E2E-CS-ASR model and found that the embedding feature representations of output tokens of code-switching languages are concentrated in disjoint clusters. % which hinder the model from switching between languages.
  %Each sub-space holds the representations of one language and are separated by a large gap which hinders the model from switching between languages.
  %We hypothesize that by matching the statistics of these sub-spaces, the code-switching capability of end-to-end CS-ASR can be highly improved.
  We hypothesize that a gap between these clusters hinders the E2E-CS-ASR from switching between languages, leading to sub-optimal performance.
  To address this issue, we propose embedding feature matching approaches based on Jensen-Shannon divergence and cosine distance constraints.
  %Indeed, we found that the end-to-end CS-ASR maps the linguistic features of each language into disjoint sub-spaces leading to sub-optimal performance.
  %which hinders the recognition of code-mixed utterances.
  %For that reason, we propose to augment the objective function of end-to-end ASR with additional constraint.
  %The imposed constraint will act as a cross-lingual signal enforcing the disjoint sub-spaces to be close to each other.
  The proposed constraints will act as a cross-lingual signal enforcing the disjoint clusters to be similar.
  %We investigated the Jensen-Shannon divergence and cosine distance based constraints. 
  %The experiment results performed on Mandarin-English language pair as a case study show that proposed method significantly improves the performance of end-to-end CS-ASR built using only monolingual data. 
  The experiment results performed on Mandarin-English code-switching language pair from the SEAME corpus demonstrate high effectiveness of the proposed method. 
\fi
\end{abstract}

\section{Introduction}
%The code-mixing is a practice of using more than one language within single discourse.
%The two common forms of code-mixing are inter-sentential and intra-sentential.
%In inter-sentential case, the language switches between utterances, while in intra-sentential the switch occurs within single utterance.
%In this work, we focus on intra-sentential code-mixing which is considered more common and challenging case.

%%What is the code-mixing?
The code-switching (CS) is a practice of using more than one language within a single discourse which poses a serious problem to many speech and language processing applications.
Recently, the end-to-end code-switching automatic speech recognition (E2E-CS-ASR) gained increasing interest where impressive improvements have been reported~\cite{zeng2018end,luo2018towards,li2019towards,Changhao2019}.
The improvements are mainly achieved for CS languages where sufficient amount of transcribed CS data is available such as Mandarin-English~\cite{DBLP:conf/interspeech/LyuTCL10}.
%and Hindi-English~\cite{DBLP:journals/corr/abs-1810-00662}.
%%What is the problem?
Unfortunately, for the vast majority of other CS languages the CS data remains too small or even non-existent.
%while monolingual data for each of the languages might be still available.

%%Who tried to solve the problem and how?
Several attempts have been made to alleviate the CS data scarcity problem.
Notably,~\cite{DBLP:journals/speech/YilmazMHL18,DBLP:conf/interspeech/GuoXXC18} used semi-supervised approaches to utilize untranscribed CS speech data.
On the other hand,~\cite{luo2018towards,li2019towards,Changhao2019} employed transfer learning techniques where additional monolingual speech corpora are either used for pre-training or joint-training. 
On the account of increased training data, these approaches achieved significant improvements.
%%What is the limitation on proposed approaches?
However, all these approaches rely on the cross-lingual signal imposed by some CS data or other linguistic resources such as a word-aligned parallel corpus.

%%How you want to solve the problem?
In this work, we aim to build an E2E-CS-ASR using only monolingual data without any form of cross-lingual resource.
The only assumption we make is an availability of monolingual speech corpus for each of the CS languages.
This setup is important and common to many low-resource CS languages, but has not received much research attention.
Besides, it will serve as a strong baseline performance that any system trained on CS data should reach. 

%We employ E2E-CS-ASR model based on hybrid CTC/Attention architecture~\cite{DBLP:conf/icassp/KimHW17}.
%While these models require hundreds of hours of transcribed training data, they eliminate the out-of-vocabulary problem and the need for handcrafted %pronunciation dictionary which greatly simplifies the ASR building pipeline~\cite{DBLP:journals/jstsp/WatanabeHKHH17}.
%In addition, the CTC module will enforce monotonic alignment between input speech and output token sequences. 

%In addition, they greatly simplify the ASR building process and don't require a strong expertise in the field which are important factors for low-resource languages.
%To the best of our knowledge, no attempt has been made to build an end-to-end CS-ASR using only monolingual data.
%To the best of our knowledge, building an E2E-CM-ASR using only monolingual data haven't been studied yet.

%However, the E2E-CS-ASR model will fail to learn language switch-points between and within utterances due to the absence of CS train data.
However, due to the absence of CS train data, the E2E-CS-ASR model will fail to learn cross-lingual relations between monolingual languages.
Consequently, the output token embeddings of monolingual languages will diverge from each other, and hence, prevent the E2E-CS-ASR model from  switching between languages.
%Consequently, the output token distributions of monolingual languages will be independent from each other.
%One of the major drawbacks of using only monolingual data to train E2E-CS-ASR is the absence of cross-lingual signal.
%As a result, the E2E-CM-ASR will fail to learn language switch-points within single utterance, leading to inter-sentential\footnote{A mode of CM where switching occurs only between utterances.} system.
%As a result, the E2E-CS-ASR will fail to learn language switch-points between and within utterances.
%Indeed, we examined the shared feature space modeled by output projection matrix in the decoder module of E2E-CM-ASR, and observe that learned representations of output tokens form two disjoint sub-spaces (see Figure~\ref{fig:no_constraint}).
%Indeed, we examined the shared output token embedding space learned by E2E-CS-ASR and observe that output token embeddings of two monolingual languages are concentrated in two disjoint clusters (see Figure~\ref{fig:no_constraint}).
Indeed, we examined the shared output token embedding space learned by E2E-CS-ASR and observed that output token embeddings of two monolingual languages are differently distributed and located apart from each other (see Figure~\ref{fig:no_constraint}).
We hypothesize that the difference between output token embedding distributions restricts the E2E-CS-ASR model from correctly recognizing CS utterances.
To address this problem, we propose to impose additional constraints which will encourage output token embeddings of monolingual languages to be similar.
Specifically, we propose to use Jensen-Shannon divergence and cosine distance based constraints.
The  former  will  enforce  output token embeddings  of monolingual languages to possess similar distributions, while the later simply brings the centroids of two distributions to be close to each other. 
%These constraints are incorporated into the objective function of E2E-CS-ASR where they will act as a cross-lingual signal source forcing the output token embeddings of monolingual languages to be similar. 
%Specifically, the imposed constraints will act as a cross-lingual signal which pulls together the two clusters.
%The imposed constraint will act as an ``unsupervised" cross-lingual signal which pulls together the two clusters.
%Specifically, the imposed constraint will forces the learned embedding vectors of two languages to be similar.
%The proposed method can be viewed as an "unsupervised" cross-lingual signal generator which typically pulls together the two sub-spaces.
%Our method typically serves as cross-lingual signal which pulls together the two separate sub-spaces. 
%For instance, we used cosine similarity metric to minimize the distance between centroids of two sub-spaces.
%In the same spirit, we also examined Jensen-Shannon divergence to minimize the distribution difference between features.
In addition, the imposed constraints will act as a regularization term to prevent overfitting. % and help to improve overall recognition accuracy.
Our method is inspired by~\cite{DBLP:conf/interspeech/KaritaWIOD18,DBLP:conf/slt/DrexlerG18} where intermediate feature representations of text and speech are forced to be close to each other.
%using generative adversarial networks and Kullback-Leibler divergence.
%We investigated Jensen-Shannon divergence and cosine distance based constraints.
We evaluated our method on Mandarin-English CS language pair from the SEAME~\cite{DBLP:conf/interspeech/LyuTCL10} corpus where we removed all CS utterances from the training data.
%As a case study we used Mandarin-English language pair from the SEAME dataset where we removed all code-mixed utterances.
%We examined our method on the state-of-the-art end-to-end CS-ASR built using recent trends such as data augmentation and word-pieces.
Experimental results demonstrate high effectiveness of the proposed method, yielding up to $4.5\%$ absolute mixed error rate improvement.
%recent trends such as data augmentation~\cite{DBLP:conf/interspeech/KoPPK15} and subword units~\cite{DBLP:conf/acl/SennrichHB16a}.

%Experiment results performed on the SEAME dataset shows that our method significantly improves the ASR performance over the strong baseline model.
%We evaluated our method on strong end-to-end ASR system[] using SEAME dataset where code-mixed utterances were removed from the training set.

The rest of the paper is organized as follows.
In Section 2, we review related works addressing the CS data scarcity problem.
In Section 3, we briefly describe the baseline E2E-CS-ASR model. 
In Section 4, we present the constrained output embeddings method.
Section 5 describes the experiment setup and discusses the obtained results.
Lastly, Section 6 concludes the paper.

\section{Related works}
%In this section, we review some of the approaches designed to deal with the code-mixed data scarcity problem.
%These approaches are mainly based on semi-supervised and transfer learning based techniques.

An early approach to build CS-ASR using only monolingual data is so-called ``multi-pass" system~\cite{DBLP:conf/icassp/LyuLCH06}.
The multi-pass system is based on traditional ASR and consists of three main steps.
First, the CS utterances are split into monolingual speech segments using the language boundary detection system.
Next, obtained segments are labeled into specific languages using the language identification system.
Lastly, labeled segments are decoded using the corresponding monolingual ASR system.
However, this approach is prone to error-propagation between different steps.
Moreover, the language boundary detection and language identification tasks are considered difficult.
%This approach is inconvenient and prone to error-propagation between different steps, not to mention that language boundary detection and language identification are difficult tasks.
%In addition to the challenging sub-tasks, this approach is complicated and prone to error-propagation between sub-tasks.
%This approach, however, is complicated and prone to error-propagation between systems.

More recently, the semi-supervised approaches have been explored to circumvent the CS data scarcity problem.
For instance,~\cite{DBLP:journals/speech/YilmazMHL18} used their best CS-ASR to transcribe a raw CS speech, the transcribed speech is then used to re-train the CS-ASR.
%It is worth to mention that~\cite{DBLP:journals/speech/YilmazMHL18} also explored multi-pass based semi-supervised technique where raw speech segments are first labeled using language identification system and then transcribed using corresponding monolingual ASR.
In a similar manner,~\cite{DBLP:conf/interspeech/GuoXXC18} employed their best CS-ASR to re-transcribe the poorly transcribed portion of the training set and then use it to re-train the model.
%achieve considerable WER improvements.
Although the semi-supervised approaches are promising, they still require CS data as well as other systems such as language identification.
%The semi-supervised approaches are promising direction for increasing CS data, however, they depends on the availability of transcribed CS data and other systems such as language identification.

In the context of end-to-end ASR models, the transfer learning techniques are widely used to alleviate the CS data scarcity problem.
For example,~\cite{luo2018towards,li2019towards} used monolingual data to pretrain the model followed by the fine-tuning with CS data.
On the other hand,~\cite{Changhao2019} used both CS and monolingual data for pre-training followed by the standard fine-tuning with the CS only data.
While being effective, the transfer learning based techniques highly rely on the CS data.

Generating synthesized CS data using only monolingual data has been also explored in~\cite{DBLP:conf/acl/ChoudhuryDBSPB18,DBLP:journals/corr/abs-1811-02356,DBLP:journals/corr/abs-1810-10254,DBLP:conf/interspeech/YilmazHL18}, however, they only address the textual data scarcity problem.

\section{Baseline E2E-CS-ASR}
\begin{figure}[t]
    \centering
    \includegraphics[width=0.8\linewidth]{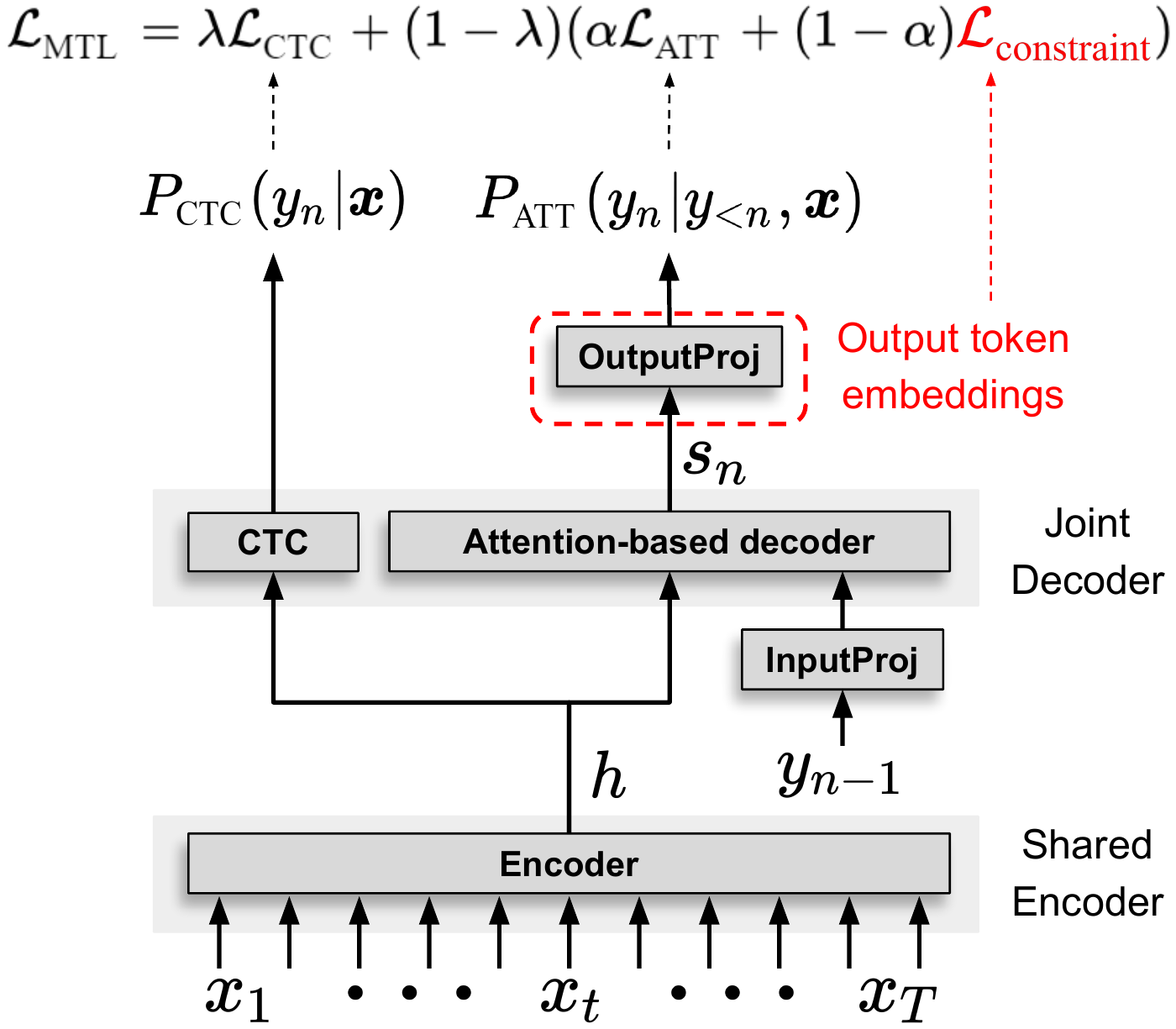}
    \caption{Hybrid CTC/Attention end-to-end ASR architecture with constrained output token embeddings. The output token embeddings are learned by the parametric matrix of linear output projection layer (OutputProj).}
    \label{fig:hybrid}
\end{figure}

%This section only focuses on the modules impacted by the proposed method.
%The detailed explanation of the other modules can be found in~\cite{DBLP:journals/jstsp/WatanabeHKHH17}.
%There are two major types of end-to-end architecture for ASR: Attention based encoder-decoder models and Connectionist Temporal  Classification (CTC) models.
%Recently, a joint  CTC/Attention architecture, integrating the advantages of both models, have been proposed~\cite{DBLP:conf/icassp/KimHW17}.
Figure~\ref{fig:hybrid} illustrates the baseline E2E-CS-ASR model based on hybrid CTC/Attention architecture~\cite{DBLP:conf/interspeech/HoriWZC17} which incorporates the advantages of both Connectionist Temporal Classification (CTC) model~\cite{DBLP:conf/icml/GravesFGS06} and attention-based encoder-decoder model~\cite{DBLP:conf/icassp/BahdanauCSBB16}.
Specifically, the CTC and attention-based decoder modules share a common encoder network and are jointly trained.

\textbf{Encoder.} The shared encoder network takes a sequence of $T$-length speech features $\boldsymbol{x}=(x_1,\dots,x_T)$ and transforms them into $L$-length high level representations $\boldsymbol{h}=(h_1,\dots,h_L)$ where $L<T$.
The encoder is modeled as a deep convolutional neural network (CNN) based on the VGG network~\cite{DBLP:journals/corr/SimonyanZ14a} followed by several bidirectional long short-term memory (BLSTM) layers.
\begin{equation}
    \boldsymbol{h}=\text{BLSTM}(\text{CNN}(\boldsymbol{x}))
    \label{eq:encoder}
\end{equation}

\textbf{CTC module.} The CTC sits on top of the encoder and computes the posterior distribution $P_{\text{\tiny CTC}}(\boldsymbol{y}|\boldsymbol{x})$ of $N$-length output token sequence $\boldsymbol{y}=(y_1,\dots,y_N)$.
The CTC loss is defined as a negative log-likelihood of the ground truth sequences $\boldsymbol{y}^*$:
\begin{equation}
    \mathcal{L}_{\text{\tiny CTC}}=-\log P_{\text{\tiny CTC}}(\boldsymbol{y}^*|\boldsymbol{x})
    \label{eq:ctc_loss}
\end{equation}

\ifx
\textbf{CTC module.} The CTC sits on top of encoder and computes the posterior distribution $P_{\text{\tiny CTC}}(\boldsymbol{y}|\boldsymbol{x})$ of $N$-length output characters sequence $\boldsymbol{y}=(y_1,\dots,y_N)$.
To compute $P_{\text{\tiny CTC}}(\boldsymbol{y}|\boldsymbol{x})$, CTC introduces framewise letter sequence with an additional ``blank" symbol $\boldsymbol{z}=(z_1,\dots,z_T)$ and factorizes $P_{\text{\tiny CTC}}(\boldsymbol{y}|\boldsymbol{x})$ using conditional independence assumption as follows:
\begin{equation}
    P_{\text{\tiny CTC}}(\boldsymbol{y}|\boldsymbol{x})\approx\sum_{\boldsymbol{z}}\prod_{t}P(z_t|z_{t-1},\boldsymbol{y})P(z_t|\boldsymbol{x})P(\boldsymbol{y})
    \label{eq:ctc}
\end{equation}
where three distribution components are: state transition probability $P(z_t|z_{t-1},\boldsymbol{y})$, framewise posterior distribution $P(z_t|\boldsymbol{x})$ and character-level language model $P(\boldsymbol{y})$.
The state transition probability enforces the monotonic alignment between speech and character sequences, and is obtained using the set of pre-defined rules (see Eq. (21) in~\cite{DBLP:journals/jstsp/WatanabeHKHH17}), whereas framewise posterior distribution is modeled as follows:
\begin{equation}
    P(z_t|\boldsymbol{x})=\text{Sofmax}(\text{Lin}(\boldsymbol{h}))
    \label{eq:frame_post}
\end{equation}
where Lin($\cdot$) is a linear projection layer with learnable matrix and bias parameters.
Lastly, the Eq.~\eqref{eq:ctc} is efficiently computed using dynamic programming.
The CTC loss is defined as a negative log-likelihood of the ground truth character sequences $\boldsymbol{y}^*$:
\begin{equation}
    \mathcal{L}_{\text{\tiny CTC}}=-\log P_{\text{\tiny CTC}}(\boldsymbol{y}^*|\boldsymbol{x})
    \label{eq:ctc_loss}
\end{equation}
\fi

\textbf{Attention-based decoder module.} %Unlike the CTC module, the attention-based decoder directly computes the $P_{\text{\tiny ATT}}(\boldsymbol{y}|\boldsymbol{x})$ based on the chain rule:
The attention-based decoder computes the probability distribution $P_{\text{\tiny ATT}}(\boldsymbol{y}|\boldsymbol{x})$ over the output token sequence $\boldsymbol{y}$ given the previously emitted tokens $y_{<n}$ and input feature sequence $\boldsymbol{x}$ using the chain rule:
\begin{align}
    \alpha_{n}&=\text{Attention}(s_{n-1},\alpha_{n-1},\boldsymbol{h}) \label{eq:decoder_att} \\
    c_{n}&=\sum^{L}_{j=1}\alpha_{n,j}h_j \\
    s_{n}&=\text{LSTM}(s_{n-1},c_{n},\text{InputProj}(y_{n-1})) \label{eq:decoder_lstm}\\
    P(y_{n}|y_{<n},\boldsymbol{x})&=\text{Softmax}(\text{OutputProj}(s_{n})) \label{eq:decoder_output_proj}\\
    P_{\text{\tiny ATT}}(\boldsymbol{y}|\boldsymbol{x})&=\prod_{n}P(y_{n}|y_{<n},\boldsymbol{x}) \label{eq:decoder}
\end{align}
where $\alpha_{n}$ is an attention weight vector produced by Attention($\cdot$) module,
$c_{n}$ is a context vector which encapsulates the information in the input speech features required to generate the next token,
$s_{n}$ is a hidden state produced by unidirectional long short-term memory (LSTM). 
% which accepts previous hidden state $s_{n-1}$, previously emitted character $y_{n-1}$ and context vector $c_{n}$.
InputProj($\cdot$) and OutputProj($\cdot$) are input and output linear projection layers with learnable matrix parameters, respectively.
The input and output learnable matrices hold input and output embedding representations of tokens, respectively. 
%$y_{<n}$ is a sequence of previously emitted characters. OutputProj($\cdot$) and InputProj($\cdot$) are output and input linear projection layers with learnable matrix parameters, repectively.
%$s_{n}$ is a hidden state produced by unidirectional long short-term memory (LSTM) which accepts previous hidden state $s_{n-1}$, previously emitted character $y_{n-1}$ and context vector $c_{n}$.
%The context vector $c_{n}$ encapsulates the information in the input speech features required to generate the next character and is produced by Attention($\cdot$) module. % based on hybrid attention mechanism~\cite{DBLP:conf/nips/ChorowskiBSCB15}.
The loss function of attention-based decoder module is computed using Eq.~\eqref{eq:decoder} as:
\begin{equation}
    \mathcal{L}_{\text{\tiny ATT}}=-\log P_{\text{\tiny ATT}}(\boldsymbol{y}^*|\boldsymbol{x})
    \label{eq:a_loss}
\end{equation}

Finally, the CTC and attention-based decoder modules are jointly trained within multi-task learning (MTL) framework as follows:
\begin{equation}
    \mathcal{L}_{\text{\tiny MTL}}=\lambda\mathcal{L}_{\text{\tiny CTC}}+(1-\lambda)\mathcal{L}_{\text{\tiny ATT}}
    \label{eq:mtl}
\end{equation}
where $\lambda$ controls the contribution of the losses.

Our proposed method will append additional constraint into the MTL framework which will mainly impact the learnable matrix parameter of OutputProj($\cdot$) layer in Eq.~\eqref{eq:decoder_output_proj} as will be explained in the following section.

\section{Constrained output embeddings}
In this work, we aim to build E2E-CS-ASR using only monolingual data.
This setup is essential for the vast majority of CS languages for which CS data is non-existent.
%However, an E2E-CS-ASR model trained on monolingual data will fail to learn language switch-points between and within utterances, due to the absence of cross-lingual signal.
However, an E2E-CS-ASR model trained on monolingual data will fail to learn language switch-points, and hence, will perform sub-optimally on input CS speech.
We investigated the E2E-CS-ASR model and found that the output token representations of monolingual languages, modeled by linear projection layer OutputProj($\cdot$), to be different and apart from each other (see Figure~\ref{fig:no_constraint}).
%However, the output token representations learned by output embedding matrix (Eq.~\eqref{eq:decoder_output_proj}) of attention-based decoder module are concentrated in two disjoint clusters (see Figure~\ref{fig:no_constraint}).    
%that would guide the model through the language switch-points.
%Consequently, the model's performance on CM test data will be sub-optimal.
%Consequently, the model will only suit to operate on inter-sentential mode.
%The further analysis of the shared feature space learned by E2E-CS-ASR model confirm that the learned output token representations, stored in the linear projection matrix of decoder module (Eq.~\eqref{eq:decoder_output_proj}), are separated into two sub-spaces (see Figure~\ref{fig:no_constraint}).
%We analyzed the E2E-CM-ASR model, and found that output projection matrix at the decoder module (Eq.~\eqref{eq:decoder_output_proj}) partitions the embeddings of output tokens into two separate sub-spaces (see Figure).
%Each sub-space corresponds to output tokens of one language.
%Each cluster holds the output token representations of one language.
We hypothesize that the difference between output token distributions of monolingual languages restricts the E2E-CS-ASR model from switching between languages.
%We suspect that the large gap between two sub-spaces hinders the model from switching between languages, and thus, leading to sub-optimal performance on the code-mixed utterances.

%To address this problem, we propose to match the statistics of two sub-spaces using additional constraint.
%To bridge the gap between these distributions, we propose to constrain output token embeddings using Jensen-Shannon divergence (JSD) and cosine distance (CD).
To reduce the discrepancy between these distributions, we propose to constrain output token embeddings using Jensen-Shannon divergence (JSD) and cosine distance (CD).
These constraints will typically act as a cross-lingual signal source which will force output token embedding representations of monolingual languages to be similar.
%The imposed constraint will typically act as a cross-lingual signal which pulls together the two sub-spaces.
%In addition, it will serve as a regularization term improving the overall performance.
%We investigate two types of constraints based on $(1)$ Jensen-Shannon divergence (JSD) and $(2)$ cosine distance (CD).
Specifically, JSD will enforce the output token embeddings of monolingual languages to possess similar distributions.
On the other hand, CD will enforce the centroids of two distributions to be close to each other.

\textbf{Jensen-Shannon divergence.} 
First, we assume that learned output token embeddings of monolingual language pair $L_1$ and $L_2$ follow a $z$-dimensional multivariate Gaussian distribution:
\begin{align}
    L_1&\sim Normal(\mu_{1},\Sigma_{1})\\
    L_2&\sim Normal(\mu_{2},\Sigma_{2})
    \label{eq:mgd}
\end{align}
The JSD between these distributions is then computed as:
\begin{align}
    \mathcal{L}_{\text{\tiny JSD}}=&tr(\Sigma^{-1}_{1}\Sigma_{2}+\Sigma_{1}\Sigma^{-1}_{2}) \nonumber\\
                                  &+(\mu_1-\mu_2)^{T}(\Sigma^{-1}_{1}+\Sigma^{-1}_{2})(\mu_1-\mu_2)-2z
    \label{eq:js}
\end{align}

Lastly, we fuse the JSD constraint with the loss function of E2E-CS-ASR using Eq.~\eqref{eq:mtl} as follows:
\begin{equation}
    \mathcal{L}_{\text{\tiny MTL}}=\lambda\mathcal{L}_{\text{\tiny CTC}}+(1-\lambda)(\alpha\mathcal{L}_{\text{\tiny ATT}}+(1-\alpha)\mathcal{L}_{\text{\tiny JSD}})
    \label{eq:mtl_augmented}
\end{equation}
where $\alpha\in[0,1]$ controls the importance of the constraint.

\textbf{Cosine distance.}
%We first compute the average vectors $C_1$ and $C_2$ corresponding to the distribution centroids of monolingual language pair $L_1$ and $L_2$ respectively.
We first compute the centroid vectors $C_1$ and $C_2$ obtained by taking the mean of all output token embeddings of monolingual language pair $L_1$ and $L_2$, respectively.
The cosine distance between two centroids is then computed as follows:
\begin{equation}
    \mathcal{L}_{\text{\tiny CD}}=1-\frac{C_1\cdot C_2}{\left\Vert C_1\right\Vert\left\Vert C_2\right\Vert}
    \label{eq:cd}
\end{equation}
The CD constraint is integrated into the loss function in a similar way as Eq.~\eqref{eq:mtl_augmented}.

%The decoder module plays a role of conditional language model, as such, the output projection layer can be interpreted as an embedding matrix holding the dense representations of output tokens.

\section{Experiment}

\subsection{Dataset}
We evaluate our method on Mandarin-English CS language pair from the SEAME~\cite{DBLP:conf/interspeech/LyuTCL10} corpus (see Table~\ref{tab:dataset}).
We used standard data splitting\footnote{\url{https://github.com/zengzp0912/SEAME-dev-set}} on par with previous works~\cite{zeng2018end,DBLP:conf/interspeech/GuoXXC18} which consists of $3$ sets: train, test$_{man}$ and test$_{eng}$.
To match the no CS data scenario, where we assume that we only possess monolingual data, 
we removed all CS utterances from the train set.
The test$_{man}$ and test$_{eng}$ sets were used for evaluation.
Both evaluation sets are gender balanced and consist of $10$ speakers, but the matrix\footnote{The dominant language into which elements from the embedded language are inserted.} language of speakers is different, i.e. Mandarin for test$_{man}$ and English for test$_{eng}$.  

\begin{table}[t]
    \caption{SEAME dataset statistics after removing the CS utterances from the train set. `Man' and `Eng' refer to Mandarin and English languages, respectively}
    \label{tab:dataset}
    \centering
    \begin{tabular}{l|c|c|c|c}
        \toprule
        \multirow{2}{*}{}   & \multicolumn{2}{|c|}{train}   & \multirow{2}{*}{test$_{man}$}   & \multirow{2}{*}{test$_{eng}$} \\ \cline{2-3}
                            & Man           & Eng           &                               & \\ \hline
        \# tokens           & ${\sim}$216k  & ${\sim}$109k  & ${\sim}$96k                   & ${\sim}$54k \\
        \# utterances       & 21,476        & 17,925        & 6,531                         & 5,321 \\
        (\# CS utterances)  & (0)           & (0)           & (4,418)                       & (2,652) \\
        Duration            & 15.8 hr       & 11.8 hr       & 7.5 hr                        & 3.9 hr \\             
        \bottomrule
    \end{tabular}
\end{table}

\ifx
\begin{table}[t]
    \caption{SEAME dataset statistics after removing the CM utterances from the train set.}
    \label{tab:dataset}
    \centering
    \begin{tabular}{l|c|c|c}
        \toprule
                            & \textbf{\#utterances} & \textbf{\#tokens} & \textbf{duration} \\ \hline
        train\_eng   & 17,925                & 109k              & 11.8 hr \\ 
        train\_man   & 21,476                & 216k              & 15.8 hr \\
        eval\_sge     & 5,321                 & 54k               & 3.9 hr \\
        eval\_man     & 6,531                 & 96k               & 7.5 hr \\
        \bottomrule
    \end{tabular}
\end{table}
\fi

\subsection{E2E-CS-ASR model configuration}
We used ESPnet toolkit~\cite{DBLP:conf/interspeech/WatanabeHKHNUSH18} to train our baseline E2E-CS-ASR model.
%The encoder module consists of $2$ initial blocks of VGG network followed by $6$ BLSTM layers each with $512$ units.
The encoder module consists of VGG network followed by $6$ BLSTM layers each with $512$ units.
The attention-based decoder module consists of a single LSTM layer with $512$ units and employs multi-headed hybrid attention mechanism~\cite{DBLP:conf/nips/ChorowskiBSCB15} with $4$ heads.
The CTC module consists of a single linear layer with $512$ units and its weight in Eq.~\eqref{eq:mtl} is set to $0.2$.
The network was optimized using Adadelta with gradient clipping.
During the decoding stage, the beam size was set to $30$.
The baseline model achieves $34.3\%$ and $46.3\%$ mixed error rates (MER)\footnote{The term ``mixed'' refers to different token units used for English (words) and Mandarin (characters).} on test$_{man}$ and test$_{sge}$ respectively, when trained on entire SEAME train set including the CS utterances.
%The mixed error rate (MER)\footnote{The term ``mixed'' refers to different atomic units used for English (words) and Mandarin (characters).} performance of the baseline E2E-CM-ASR model trained on entire SEAME corpus, including the code-mixed utterances, is $34.3\%$ and $46.3\%$ on test$_{man}$ and test$_{sge}$ sets, respectively.
%is shown in the first row of Table~\ref{tab:results}.
%The first three rows in Table~\ref{tab:results} show mixed error rate (MER)\footnote{The term ``mixed'' refers to different atomic units used for English (words) and Mandarin (characters).} performance of the baseline E2E-CM-ASR model.

\subsection{Results and analysis}
The experiment results are shown in Table~\ref{tab:results}.
We split the test sets into monolingual and CS utterances to analyze the impact of the proposed method on each of them.
We first report the MER performance of a conventional ASR model built using Kaldi toolkit~\cite{Povey_ASRU2011} (row 1), the model specifications can be found in~\cite{DBLP:conf/interspeech/GuoXXC18}.
%on monolingual data and handcrafted pronunciation dictionary (row $2$). 
The MER performance of the baseline E2E-CS-ASR model is shown in the second row.
We followed the recent trends~\cite{zeng2018end,luo2018towards,Changhao2019} to obtain a much stronger baseline model.
Specifically, we applied speed perturbation (SP) based data augmentation technique~\cite{DBLP:conf/interspeech/KoPPK15} and used byte pair encoding (BPE) based subword units~\cite{DBLP:conf/acl/SennrichHB16a} to balance Mandarin and English tokens (rows $3$ and $4$).
We tried different vocabulary sizes for BPE and found $4k$ units to work best in our case.
%We tried different vocabulary sizes for BPE and found $4k$ units to work best in our case, resulting in much stronger baseline model.
%recent trends such as data augmentation~\cite{DBLP:conf/interspeech/KoPPK15} and subword units~\cite{DBLP:conf/acl/SennrichHB16a}
%Label smoothing is applied by weighting the ground truth with 0.95, and uniformly distributing the remaining probability mass among other tokens[].
%During the decoding, we use beam search algorithm with a beam size of 30.

\begin{table}[h]
    \caption{The MER (\%) performance of different ASR models built using monolingual data. The test sets are further split into monolingual (mono) and code-switching (CS) utterances}
    \footnotesize
    \label{tab:results}
    \centering
    \begin{tabular}{c|l|c|c|c|c|c|c}
        \toprule
        \multirow{3}{*}{No.}    & \multirow{3}{*}{Model}            & \multicolumn{3}{|c}{test$_{man}$}             & \multicolumn{3}{|c}{test$_{eng}$}         \\\cline{3-8}
                                &                                   & \scell{mono\\utts.} & \scell{CS\\utts.} & all & \scell{mono\\utts.} & \scell{CS\\utts.}& all \\\hline
        %E2E-CM-ASR                          & \multirow{2}{*}{-} & \multirow{2}{*}{-} & \multirow{2}{*}{34.3} & \multirow{2}{*}{-} & \multirow{2}{*}{-} &  \multirow{2}{*}{46.3} \\
        %(entire SEAME)                  &               &               &               &               &               &      \\\hline
        1   & Kaldi                     & -             & -             & 39.1          & -             & -             & 45.2 \\\hline
        2   & Baseline                  & 57.7          & 73.3          & 70.6          & 73.7          & 80.6          & 78.3 \\
        3   & + SP                      & 39.4          & 56.0          & 53.2          & 54.2          & 65.9          & 62.2 \\
        4   & \ \ \ + BPE               & 38.1          & 51.8          & 49.5          & 52.9          & 61.4          & 58.9 \\
        5   & \ \ \ \ \ \ + CD          & 34.4          & 49.0          & 46.3          & \textbf{47.2} & 58.5          & 55.1 \\
        6   & \ \ \ \ \ \ + JSD         & 34.9          & 48.8          & 46.3          & 47.8          & 57.6          & 54.6 \\
        7   & \ \ \ \ \ \ \ \ \  + CD   & \textbf{34.0} & \textbf{48.1} & \textbf{45.6} & \textbf{47.2} & \textbf{57.4} & \textbf{54.4} \\
        \bottomrule
    \end{tabular}
\end{table}

\ifx
\begin{table}[hb]
    \caption{The MER performance of different end-to-end code-mixed ASR models.}
    \footnotesize
    \label{tab:baseline_results}
    \centering
    \begin{tabular}{l|c|c|c|c}
        \toprule
        \multirow{2}{*}{System}   & \multirow{2}{*}{Data aug.}& \multirow{2}{*}{Subword}  & \multicolumn{2}{|c}{MER (\%)}     \\ \cline{4-5}
                            &                           &                           & test$_{man}$  & test$_{eng}$ \\ \hline
        E2E                 & No                        & No                        & 70.6          & 78.3 \\
        E2E-SP              & Yes                       & No                        & 53.2          & 62.2 \\
        E2E-SP-BPE4k        & Yes                       & Yes                       & \textbf{49.5} & \textbf{58.9} \\
        \bottomrule
    \end{tabular}
\end{table}
\fi

The performance of models employing proposed CD and JSD constraints are shown in rows $5$ and $6$, the interpolation weights for CD and JSD are set to $0.9$ and $0.97$, respectively.
Both constraints gain considerable MER improvements.
Notably, we found that CD constraint is more effective on monolingual utterances, whereas JSD constraint is more effective on CS utterances.
To complement the advantages of both constraints, we combined them as follows:
\begin{align}
    \mathcal{L}_{\text{\tiny MTL}}=&\lambda\mathcal{L}_{\text{\tiny CTC}}+(1-\lambda)(\alpha\mathcal{L}_{\text{\tiny ATT}} + \nonumber \\                                           &(1-\alpha)(\beta\mathcal{L}_{\text{\tiny JSD}}+(1-\beta)\mathcal{L}_{\text{\tiny CD}}))
    \label{eq:mtl_augmented2}
\end{align}
where $\alpha$ and $\beta$ are set to $0.05$ and $0.9$, respectively.
The combination of two constraints significantly improves the MER over the strong baseline model by $3.9\%$ and $4.5\%$ on test$_{man}$ and test$_{eng}$, respectively (row $7$).
%Note that proposed method uses neither additional linguistic resources nor external systems such as language identification.
These results suggest that the proposed method of constraining the output token embeddings is effective. 

%While aforementioned techniques considerably reduce the gap between traditional and end-to-end ASR models, the traditional ASR still performs better. %Thus, in future work, we would like to focus more on bridging the gap further and test the proposed method for tasks with larger amount of monolingual data.
%In future work, we would like to test the proposed method for tasks with larger amount of monolingual data. 
%Thus, it is natural to combine both constraints after which we achieved further MER improvements.

\subsubsection{Changing the interpolation weight}
\begin{figure}[h]
    \centering
    \includegraphics[width=\linewidth]{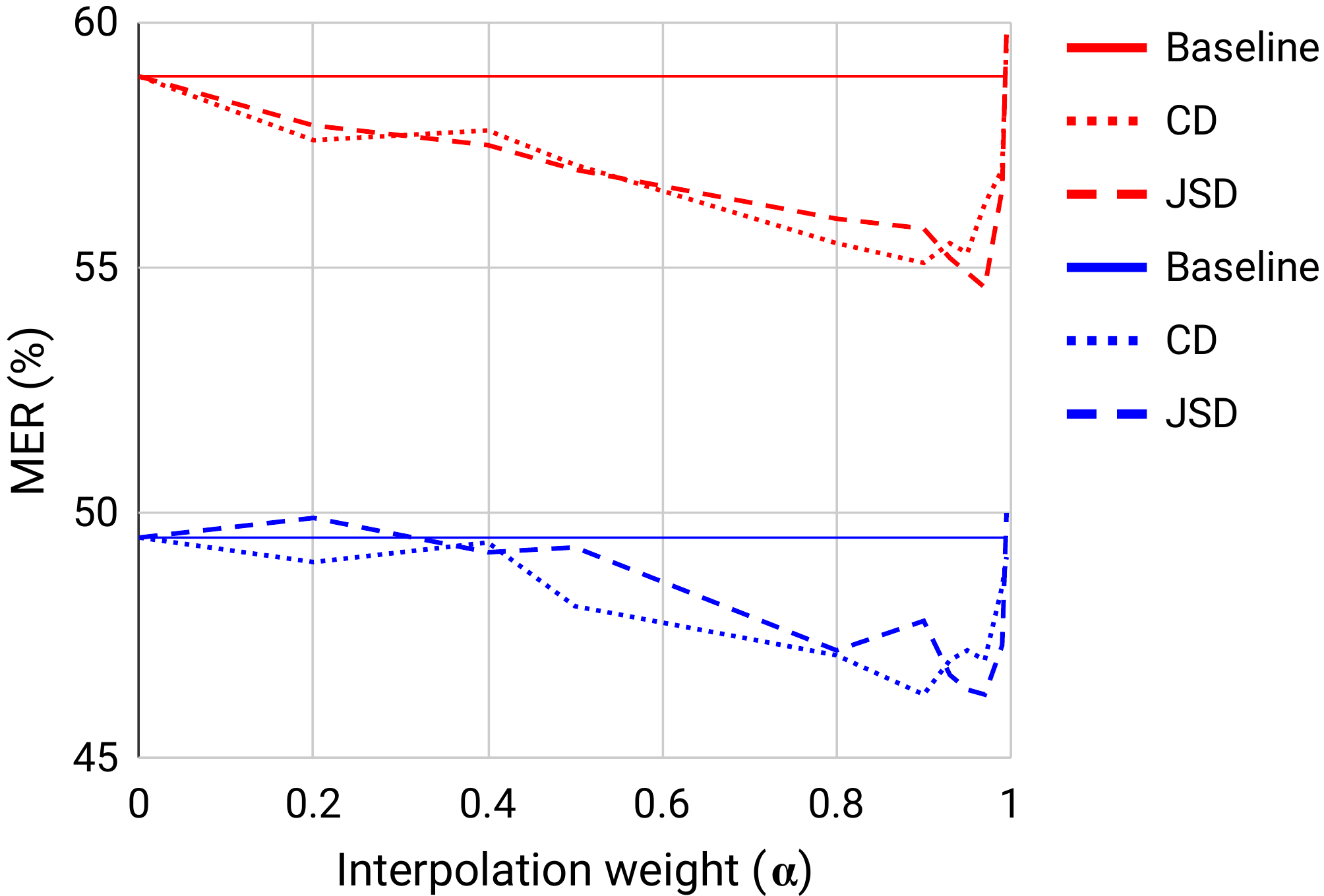}
    \caption{The impact of CD and JSD constraint interpolation weights on MER performance for test$_{eng}$ (red$/$top) and test$_{man}$ (blue$/$bottom) sets.}
    \label{fig:weights}
\end{figure}

We repeat the experiment with different interpolation weights for CD and JSD constraints (hyperparameter $\alpha$ in Eq.~\eqref{eq:mtl_augmented}) to investigate its effect on MER performance.
Figure~\ref{fig:weights} shows that the proposed method consistently improves the MER over the strong baseline model with SP and BPE.
The best results are achieved for interpolation weights in range $0.8$-$0.99$.

\begin{figure}[h]
    \centering
    \begin{subfigure}[b]{0.233\textwidth}
        \includegraphics[width=\textwidth]{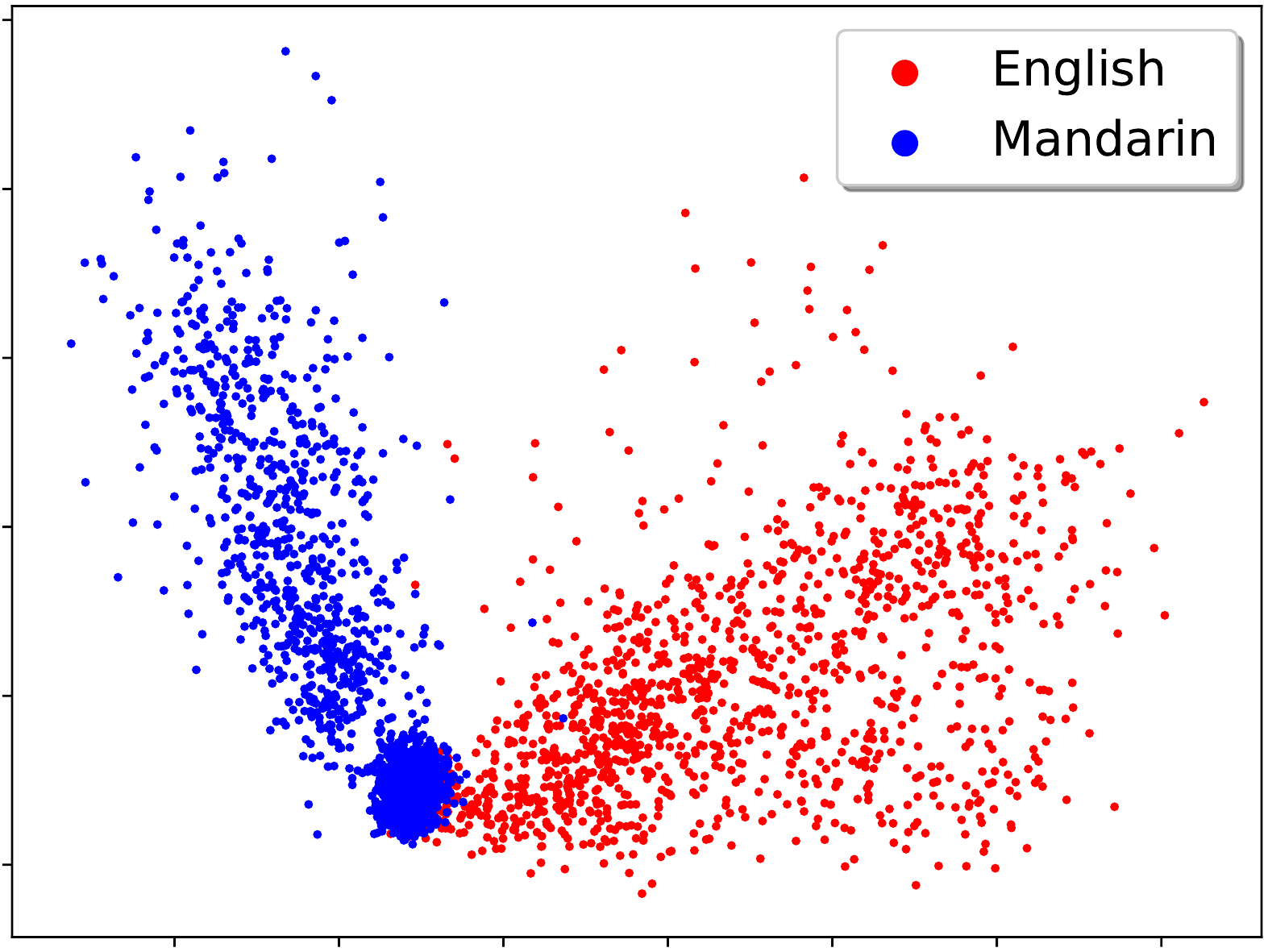}
        \vspace{-0.5cm}
        \caption{No constraint}
        \vspace{0.2cm}
        \label{fig:no_constraint}
    \end{subfigure}
    \begin{subfigure}[b]{0.233\textwidth}
        \includegraphics[width=\textwidth]{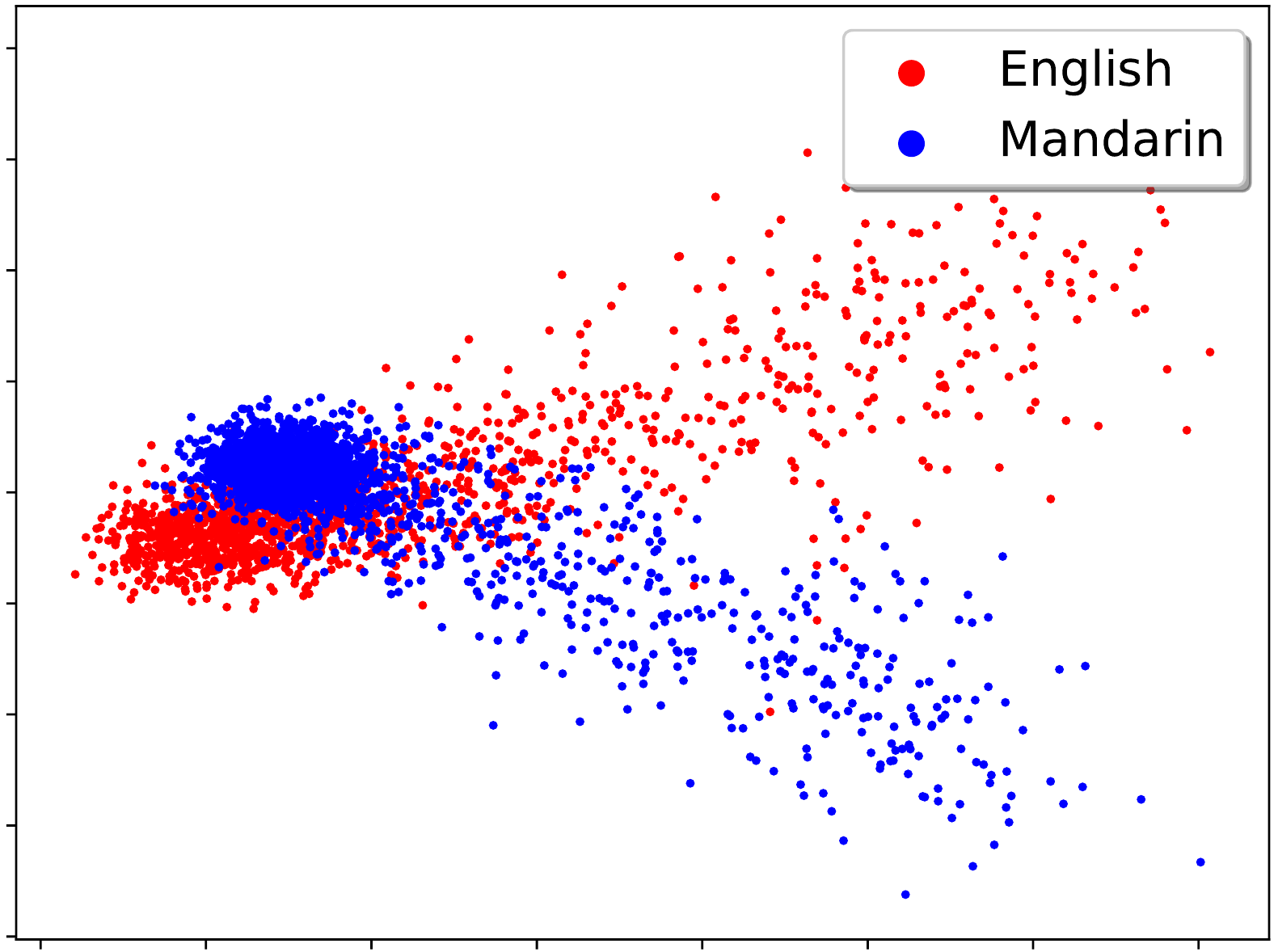}
        \vspace{-0.5cm}
        \caption{CD constraint}
        \vspace{0.2cm}
        \label{fig:cd_constraint}
    \end{subfigure}
   \begin{subfigure}[b]{0.233\textwidth}
        \includegraphics[width=\textwidth]{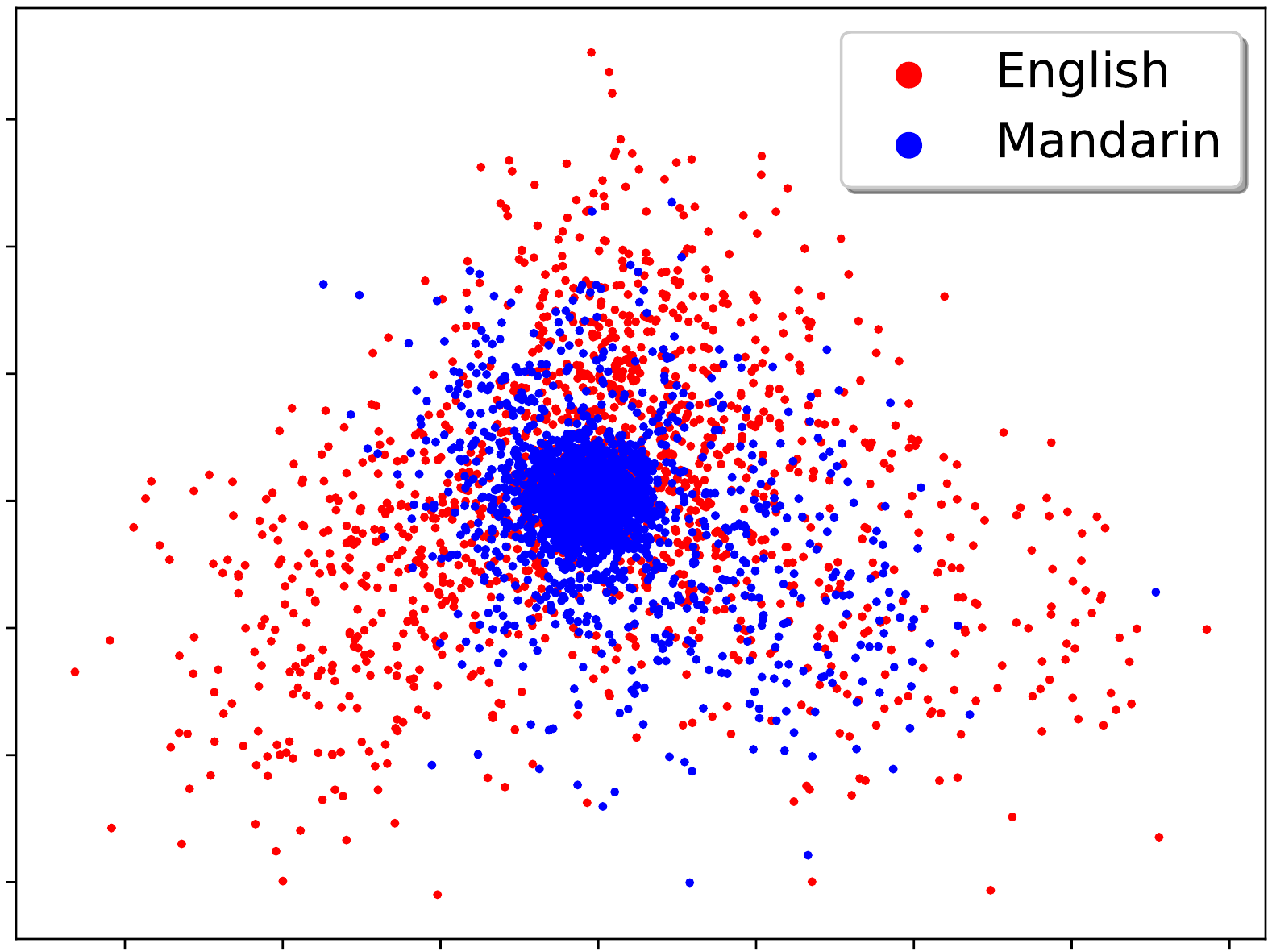}
        \vspace{-0.5cm}
        \caption{JSD constraint}
        \label{fig:js_constraint}
    \end{subfigure}
    \begin{subfigure}[b]{0.233\textwidth}
        \includegraphics[width=\textwidth]{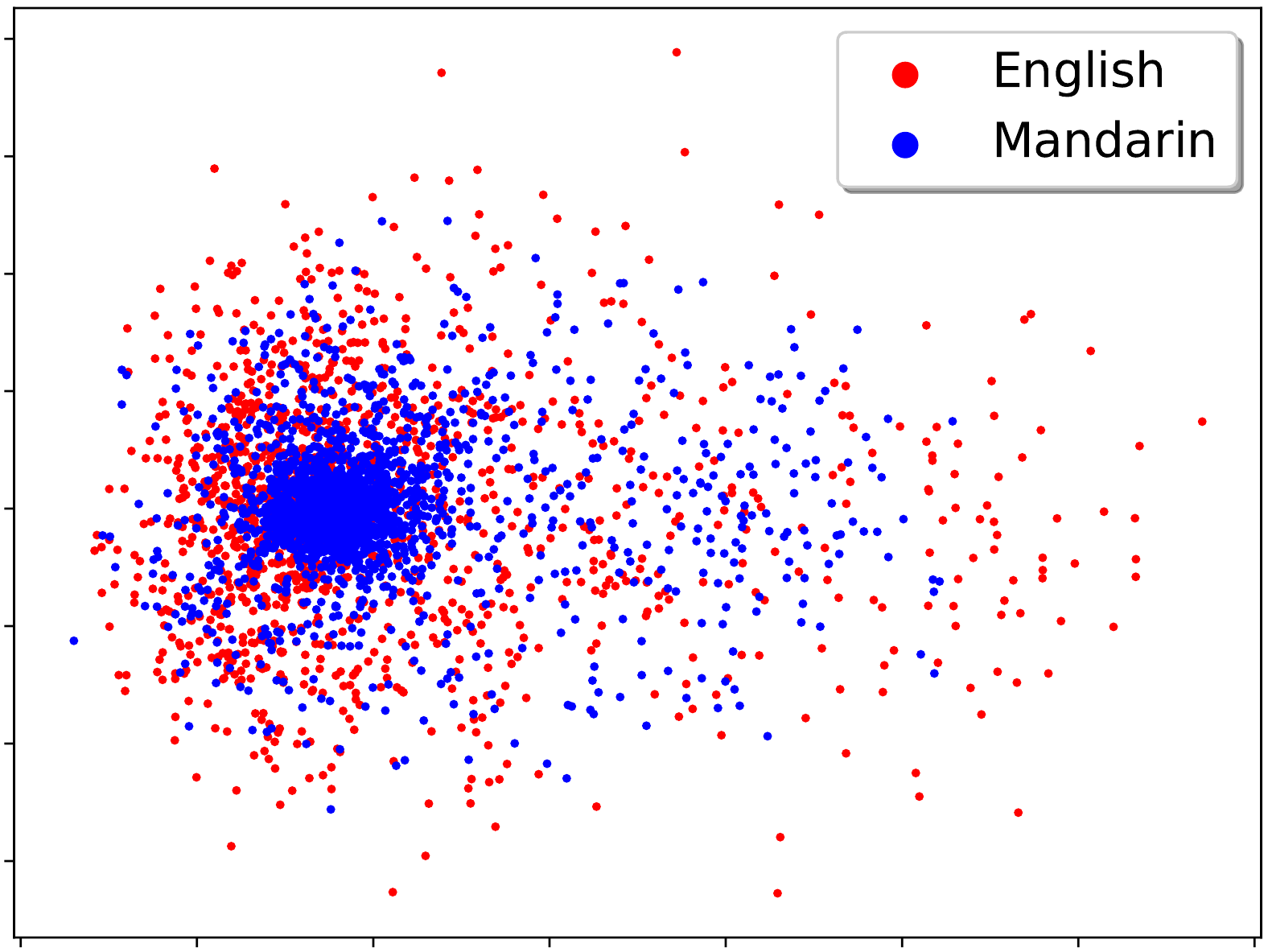}
        \vspace{-0.5cm}
        \caption{CD \& JSD constraints}
        \label{fig:cd_js_constraint}
    \end{subfigure}
    \caption{PCA visualization of shared output token embedding space without (a) and with (b,c,d) proposed constraints.}
    \label{fig:subspaces}
\end{figure}

\subsubsection{Visualization of shared output token embedding space}
To gain insights from the effects of the proposed method on the shared output embedding space, we visualize it using dimensionality reduction technique based on the principal component analysis (PCA).
Figure~\ref{fig:subspaces} shows the shared output embedding space without (\ref{fig:no_constraint}) and with (\ref{fig:cd_constraint}, \ref{fig:js_constraint}, \ref{fig:cd_js_constraint}) proposed constraints.
Note that the learned output token embeddings of monolingual languages strongly diverge from each other when proposed constraints are not employed.
Visualization of the shared output embedding space confirms that our method is effective at binding the output token embeddings of monolingual languages.
%These visualization results support our claim that imposed constraints will bridge the gap between two sub-spaces.

\subsubsection{Applying language model}
%To examine whether proposed constraints are complementary with language models (LM), we employed LMs during the decoding and rescoring stages (see Table~\ref{tab:lm}).
The state-of-the-art results in ASR are usually obtained by employing a language model (LM).
To examine whether proposed constraints are complementary with LM, we employed LM during the decoding stage.
In this experiment, we tried different LM interpolation weights changed with a step size of $0.025$ and report the best results (see Table~\ref{tab:lm}).
%The LMs were trained on monolingual transcripts of train set and applied to our best model employing both JSD and CD constraints.
%The LM was trained on monolingual transcripts of train set as a single layer LSTM with $512$ units, it was integrated using shallow fusion technique~\cite{DBLP:conf/slt/ToshniwalKCWSL18}.
The LM was trained on the entire SEAME train set, including CS utterances, as a single layer LSTM with $512$ units and was integrated using shallow fusion technique~\cite{DBLP:conf/slt/ToshniwalKCWSL18}.
%In the decoding stage, we used shallow fusion technique~\cite{DBLP:conf/slt/ToshniwalKCWSL18} to integrate subword-level recurrent LSTM LM~\cite{DBLP:conf/interspeech/SundermeyerSN12}.
%During the rescoring stage, we examined Kneser-Ney smoothed $3$-gram LM and recurrent LSTM LM, both LMs are mixed word-character level and used to rescore $50$-best hypotheses.
Obtained MER improvements show that proposed constraints and LM complement each other.
Moreover, the proposed method benefits from the LM more than the strong baseline model.

\begin{table}[h]
    \caption{The MER performance after applying the language model during the decoding stages}
    \label{tab:lm}
    \centering
    \begin{tabular}{c|c|c|c}
        \toprule
        \multirow{2}{*}{Model}  & \multirow{2}{*}{Decoder LM}   & \multicolumn{2}{|c}{MER (\%)}\\ \cline{3-4}
                                    &                               & test$_{man}$  & test$_{eng}$ \\ \hline
        \multirow{2}{*}{Baseline}   & No                            & 49.5          & 58.9 \\
                                    & Yes                           & 49.0          & 58.6 \\ \hline
        \multirow{2}{*}{Baseline + CD \& JSD} & No                  & 45.6          & 54.4 \\
                                    & Yes                           & \textbf{45.0} & \textbf{53.7} \\
%        LSTM (subword)              & 3-gram $+$ LSTM (hybrid)      &               &  \\
        \bottomrule
    \end{tabular}
\end{table}

\section{Conclusions}
%In this work, we attempt to build end-to-end code-mixed ASR using only monolingual data for low-resource language pairs.
%Due to the absence of cross-lingual signal, the ASR will fail to learn language switch-points.
In this work, we proposed a method to train improved E2E-CS-ASR model using only monolingual data.
Specifically, our method constrains the output token embeddings of monolingual languages to force them to be similar, and hence, enable E2E-CS-ASR to easily switch between languages.
We examined Jensen-Shannon divergence and cosine distance based constraints which are incorporated into the objective function of the E2E-CS-ASR.
%In particular, we examined cosine distance and Jensen-Shannon divergence based constraints.
%The former one is used to enforce learned embeddings of monolingual languages to possess similar distributions, while the later one is used to pull together centroids of output token embeddings.
We evaluated the proposed method on Mandarin-English CS language pair from the SEAME corpus where CS utterances were removed from the train set.
The proposed method outperforms the strong baseline model by a large margin, i.e. absolute $3.9\%$ and $4.5\%$ MER improvements on test$_{man}$ and test$_{eng}$, respectively.
The visualization of the shared output embedding space confirms the effectiveness of the proposed method. 
In addition, our method is complementary with the language model where further MER improvement is achieved.
Importantly, all these improvements are achieved without using any additional linguistic resources such as word-aligned parallel corpus or language identification system.
%We believe that proposed method can be easily adapted to other scenarios and can be applied to other tasks where two or more independent feature types are involved.
%We believe that the proposed method can be easily adapted to other scenarios and benefit other tasks where two or more feature types are involved.
We believe that the proposed method can be easily adapted to other scenarios and benefit other CS language pairs.

For the future work, we plan to test the proposed method on scenarios with a larger amount of monolingual data and examine its effectiveness on E2E-CS-ASR models trained using CS data.
We also plan to study the effects of the proposed method in transfer learning approach where it will be used to pre-train the model with external monolingual data.
%In addition, we plan to examine its effectiveness on E2E-CS-ASR models trained using CS data.
%For the future work, we plan to examine the effectiveness of the proposed method on E2E-CS-ASR models trained using CS data.

\ifx
We believe that proposed method of matching embedding feature representations of output tokens of CS languages can be easily adapted to other scenarios and benefit other CS language processing applications.
For the future work, we plan to test the proposed method on scenarios with larger amount of monolingual data and explore efficient ways to further improve MER performance using CS text or speech only data.
\fi

\section{Acknowledgements}
This work is supported by the project of Alibaba-NTU Singapore Joint Research Institute.

\bibliographystyle{IEEEtran}

\bibliography{mybib}

% \begin{thebibliography}{9}
% \bibitem[1]{Davis80-COP}
%   S.\ B.\ Davis and P.\ Mermelstein,
%   ``Comparison of parametric representation for monosyllabic word recognition in continuously spoken sentences,''
%   \textit{IEEE Transactions on Acoustics, Speech and Signal Processing}, vol.~28, no.~4, pp.~357--366, 1980.
% \bibitem[2]{Rabiner89-ATO}
%   L.\ R.\ Rabiner,
%   ``A tutorial on hidden Markov models and selected applications in speech recognition,''
%   \textit{Proceedings of the IEEE}, vol.~77, no.~2, pp.~257-286, 1989.
% \bibitem[3]{Hastie09-TEO}
%   T.\ Hastie, R.\ Tibshirani, and J.\ Friedman,
%   \textit{The Elements of Statistical Learning -- Data Mining, Inference, and Prediction}.
%   New York: Springer, 2009.
% \bibitem[4]{YourName17-XXX}
%   F.\ Lastname1, F.\ Lastname2, and F.\ Lastname3,
%   ``Title of your INTERSPEECH 2019 publication,''
%   in \textit{Interspeech 2019 -- 20\textsuperscript{th} Annual Conference of the International Speech Communication Association, September 15-19, Graz, Austria, Proceedings, Proceedings}, 2019, pp.~100--104.
% \end{thebibliography}

\end{document}